%% file: main.tex
\documentclass[10pt,twocolumn,letterpaper]{article}

\usepackage{cvpr}              
\input{preamble}

\definecolor{cvprblue}{rgb}{0.21,0.49,0.74}
\usepackage[pagebackref,breaklinks,colorlinks,citecolor=cvprblue]{hyperref}

\def\paperID{*****} 

\title{Density Estimation and Crowd Counting}

\author{Shantanu Todmal\\
{\tt\small stodmal@umass.edu}
\and
Rakshith Venkatesh\\
{\tt\small rakshith@umass.edu}
\and
Balachandra Devarangadi Sunil\\
{\tt\small Bdevarangadi@umass.edu}
}
\begin{document}
\maketitle
\input{sec/0_abstract}    
\input{sec/1_intro}

\input{sec/2_formatting}

\input{sec/3_finalcopy}

\end{document}

%% file: preamble.tex
\usepackage[dvipsnames]{xcolor}

\usepackage{tabularx}


%% file: sec/0_abstract.tex
\begin{abstract}
This study enhances a crowd density estimation algorithm originally designed for image-based analysis by adapting it for video-based scenarios. The proposed method integrates a denoising probabilistic model that utilizes diffusion processes to generate high-quality crowd density maps. To improve accuracy, narrow Gaussian kernels are employed, and multiple density map outputs are generated. A regression branch is incorporated into the model for precise feature extraction, while a consolidation mechanism combines these maps based on similarity scores to produce a robust final result. An event-driven sampling technique, utilizing the Farneback optical flow algorithm, is introduced to selectively capture frames showing significant crowd movements, reducing computational load and storage by focusing on critical crowd dynamics. Through qualitative and quantitative evaluations, including overlay plots and Mean Absolute Error (MAE), the model demonstrates its ability to effectively capture crowd dynamics in both dense and sparse settings. The efficiency of the sampling method is further assessed, showcasing its capability to decrease frame counts while maintaining essential crowd events. By addressing the temporal challenges unique to video analysis, this work offers a scalable and efficient framework for real-time crowd monitoring in applications such as public safety, disaster response, and event management.
\end{abstract}

%% file: sec/1_intro.tex
\section{Introduction}
\label{sec:intro}

Crowd density estimation and counting are fundamental tasks for ensuring public safety, effective event management, and disaster response. These systems are crucial in high-stakes scenarios involving large gatherings, such as concerts, sporting events, religious ceremonies, and public protests, where overcrowding can lead to stampedes, panic, or other emergencies. Real-time crowd monitoring allows authorities to preemptively manage such risks, ensuring safety and optimizing the allocation of resources. With the rise of smart city technologies and surveillance systems, the demand for accurate and efficient crowd monitoring has never been higher.

Traditional methods for crowd density estimation have primarily relied on image-based analysis techniques, which, while effective in controlled scenarios, struggle to adapt to real-world complexities. These methods typically focus on static scenes and are limited in their ability to capture dynamic interactions, temporal changes, and crowd flows over time. In contrast, video-based analysis offers the potential to capture both spatial and temporal dynamics, providing a richer understanding of crowd behavior. However, transitioning from static image-based methods to video analysis introduces new challenges, including handling vast amounts of data, reducing temporal redundancy, and ensuring computational efficiency without compromising accuracy.

Diffusion models, widely recognized for their robustness in noise reduction and density estimation, provide a strong foundation for tackling these challenges. These models excel in generating high-fidelity density maps, even in densely populated or occluded environments, by iteratively denoising input data to produce accurate representations of crowd distributions. While diffusion models have been successful in static image-based crowd analysis, their direct application to video requires overcoming specific challenges such as temporal redundancy, increased computational load, and the need to adapt to dynamic crowd behaviors.

To address these challenges, we propose an innovative extension of diffusion-based models for video analysis. A key novelty of our approach lies in the integration of an event-driven sampling mechanism based on Farneback optical flow. This method selectively processes frames that exhibit significant changes in crowd movement, effectively filtering out redundant frames while preserving critical information. By allocating computational resources only to frames that capture dynamic changes, our approach optimizes both accuracy and efficiency. Initial experiments demonstrate that this event-driven sampling reduces the number of processed frames by up to 80\%, enabling real-time processing even in resource-constrained settings.

Our methodology is tested on the ShanghaiTech dataset, a widely recognized benchmark for crowd density estimation. This dataset encompasses diverse scenarios, including sparse and dense crowd settings, providing a robust framework for evaluating the adaptability and accuracy of our approach. By demonstrating strong performance across a range of conditions, our study highlights the scalability of the proposed method and its potential to address real-world challenges in crowd monitoring.

This research contributes to the broader field of computer vision by bridging the gap between static image-based methods and video-based crowd monitoring systems. By combining the strengths of diffusion models with advanced sampling techniques, we provide a scalable and efficient solution for real-time crowd density estimation. Our work not only addresses the technical challenges of video analysis but also sets the stage for future advancements in smart surveillance, public safety, and dynamic crowd behavior modeling.

%% file: sec/2_formatting.tex
\section{Related Work and Literature Review}
\label{sec:formatting}

Crowd density estimation and counting have been extensively studied over the years, with methods evolving from traditional hand-crafted feature-based approaches to deep learning-driven models. Traditional methods, including detection-based and regression-based techniques, have shown significant limitations in dense and occluded environments. Detection-based methods rely on identifying individual objects within a scene, which becomes infeasible in high-density scenarios due to overlaps and occlusions. Regression-based approaches, while useful for global crowd estimation, often fail to capture spatial distributions or adapt to varying densities effectively.

The advent of deep learning, particularly Convolutional Neural Networks (CNNs), has revolutionized the field by enabling models to learn features directly from data, bypassing the need for manual feature engineering. CNN-based models have demonstrated superior performance in handling the complexities of crowd scenes, especially in high-density and occluded settings. This shift has paved the way for innovative architectures and techniques that address the challenges of scale variation, perspective distortion, and background clutter.

Our work builds upon several key methodologies, leveraging \href{https://openaccess.thecvf.com/content/CVPR2024/papers/Ranasinghe_CrowdDiff_Multi-hypothesis_Crowd_Density_Estimation_using_Diffusion_Models_CVPR_2024_paper.pdf}{CrowdDiff (2024)} for its unique diffusion-based density estimation framework. This framework is further extended with advanced sampling techniques to enhance computational efficiency and relevance in dense, dynamic crowd settings. Below, we outline some of the foundational methods and state-of-the-art approaches that inform our work:

\begin{itemize}
    \item \textbf{CrowdDiff (2024)}: Introduced by Ranasinghe et al., CrowdDiff utilizes diffusion models for generating crowd density maps. Unlike traditional CNNs, it employs a reverse diffusion process to mitigate background noise and produce high-fidelity density maps. A key innovation is its multi-hypothesis strategy, which generates multiple realizations of density maps, capturing variations in crowd distributions. This approach enhances robustness and accuracy in challenging scenarios, particularly in highly congested and occluded environments.
    
    \item \textbf{CSRNet (2018)}: Proposed by Li et al., CSRNet improves on earlier multi-column architectures by employing dilated convolutions, which enlarge the receptive field without resorting to pooling layers. This allows the network to capture fine-grained details in highly congested scenes while maintaining spatial resolution. CSRNet has been widely recognized for its simplicity and effectiveness in generating accurate density maps in complex environments.
    
    \item \textbf{MCNN (2016)}: As one of the pioneering CNN-based approaches, MCNN introduced the concept of multi-scale convolutional layers to handle variations in crowd densities and perspectives. By utilizing different receptive field sizes, MCNN adapts to diverse density distributions and resolutions within a single image. However, its reliance on fixed receptive field sizes and lack of explicit mechanisms for handling occlusions limits its performance in extremely dense scenes.
    
    \item \textbf{Survey by Gao et al. (2020)}: This comprehensive survey reviews over 220 CNN-based methods for crowd counting and density estimation, highlighting the evolution from simple detection-based methods to sophisticated deep learning architectures. It provides critical insights into state-of-the-art models, their strengths and limitations, and open challenges, such as handling extreme occlusions and improving computational efficiency for real-time applications.
\end{itemize}

Beyond these baseline methods, recent advancements have explored the integration of attention mechanisms, generative models, and stochastic sampling strategies to further improve accuracy and robustness. For instance: \\
- \textbf{Attention Mechanisms}: Methods like SACAN (Scale-Adaptive Crowd Attention Network) incorporate attention layers to dynamically focus on relevant crowd regions, improving accuracy in scenes with scale variations and cluttered backgrounds. \\
- \textbf{Transformer-based Models}: Vision Transformers (ViT) and related architectures have demonstrated significant potential in capturing long-range dependencies in crowd scenes. These models use self-attention mechanisms to process crowd images holistically, addressing scale variations and occlusions more effectively than traditional CNN-based approaches. \\
- \textbf{Generative Models}: GAN-based approaches (e.g., SSIMNet) have demonstrated potential for generating sharper density maps, leveraging adversarial training to enhance structural similarity with ground truth maps. \\
- \textbf{Stochastic Sampling}: Techniques such as those employed in CrowdDiff use probabilistic sampling to generate multiple density map realizations, enabling models to better capture variability in crowd distributions.

Despite these advancements, most methods are primarily designed for static images, with limited attention given to the temporal dynamics of crowd scenes in videos. Our work addresses this gap by extending image-based methodologies to video, introducing event-driven sampling to optimize computational efficiency while preserving the temporal and spatial fidelity of crowd dynamics.

In summary, our approach integrates the strengths of diffusion-based models like CrowdDiff with novel sampling techniques, advancing the state of the art in crowd density estimation and providing a scalable solution for real-time applications.

%% file: sec/3_finalcopy.tex
\section{Technical Approach}
\label{sec:technical}

\begin{figure*}[ht]
    \centering
    \includegraphics[width=1\textwidth]{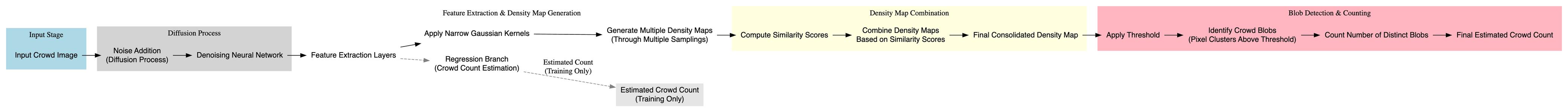} 
    \caption{Flow of paper's  approach} 
    \label{fig:p_flow} 
\end{figure*}

We plan to utilize the existing algorithm from this \href{https://openaccess.thecvf.com/content/CVPR2024/papers/Ranasinghe_CrowdDiff_Multi-hypothesis_Crowd_Density_Estimation_using_Diffusion_Models_CVPR_2024_paper.pdf}{research paper} and extend it for videos.
It is summarized as below and represented in Fig \ref{fig:p_flow}:
\begin{itemize}
    \item This approach uses a denoising probabilistic model to generate accurate crowd density maps from input images. Diffusion models gradually add noise to input images, and a denoising model learns to reverse the process.
\item It uses narrow Gaussian kernels instead of broad Gaussian kernels , which helps in capturing finer details in crowd density maps.
\item The method generates multiple density maps for the same crowd image, as it allows capturing different aspects of the crowd leading to more robust prediction.
\item During training, the model includes a regression branch that helps it learn better features for estimating crowd counts, improving the model's ability to predict crowd density accurately.
\item Finally, the method combines multiple density maps based on similarity scores to produce a final, consolidated crowd density map.
\item The pixel values in the density map that are above threshold are considered as "crowd blobs" or areas where people are present, while those below the threshold are ignored. This helps in distinguishing actual crowd areas from background noise.
\item Once the threshold is applied, the model counts the number of distinct blobs or clusters of pixels that represent people in the density map. Each blob corresponds to a group of pixels that indicates the presence of individuals. 
\item The total number of counted blobs gives the estimated crowd count for the image
\end{itemize}

\begin{figure}[htbp]
    \centering
    \includegraphics[width=0.5\textwidth]{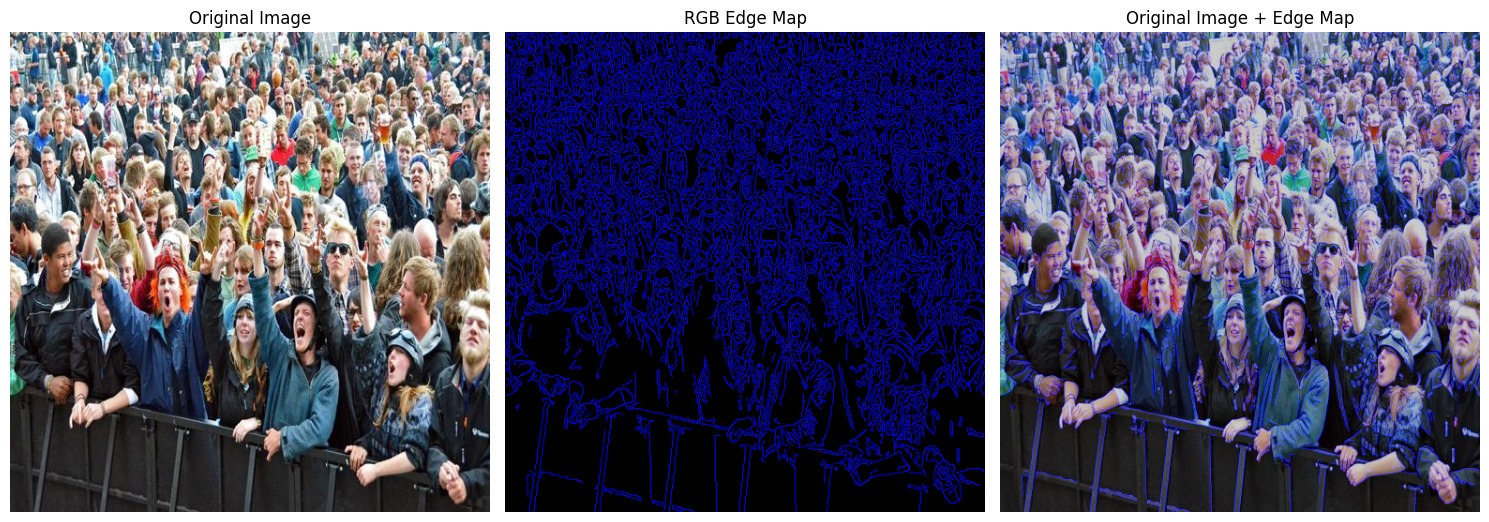} 
    \caption{Combining Edge Detection} 
    \label{fig:edge} 
\end{figure}

To extend the algorithm to video, a major challenge is reducing inference time for each frame. The current approach employs a sliding window technique to infer each crop individually and then combines the results to generate the final density map. While effective, this method is time-intensive and demands substantial GPU resources. To address this, we propose resizing the entire image to a smaller resolution of 256×256 and processing it as a whole, rather than using sliding window crops. This significantly reduces the inference time for a single image. However, down scaling the image resolution can potentially impact accuracy. To mitigate this, we enhance edge sharpness and clarity by overlaying an edge detection layout on the image, as shown in Fig \ref{fig:edge}, thereby compensating for the loss of detail.

For evaluation of the above approach, the main output will be crowd density maps, which is visually compared to the ground truth maps to verify how well the model captures crowd details, especially in dense and sparse regions. Overlay plots will be our choice for Qualitative analysis, to overlay the generated density maps on the original crowd images, inspecting whether the crowd masses align correctly with people in the image. Quantitatively,
Mean Absolute Error (MAE) will be used to measure how close the predicted crowd count is to the actual count, providing a direct measure of counting error.

\begin{figure}[htbp]
    \centering
    \includegraphics[width=0.5\textwidth]{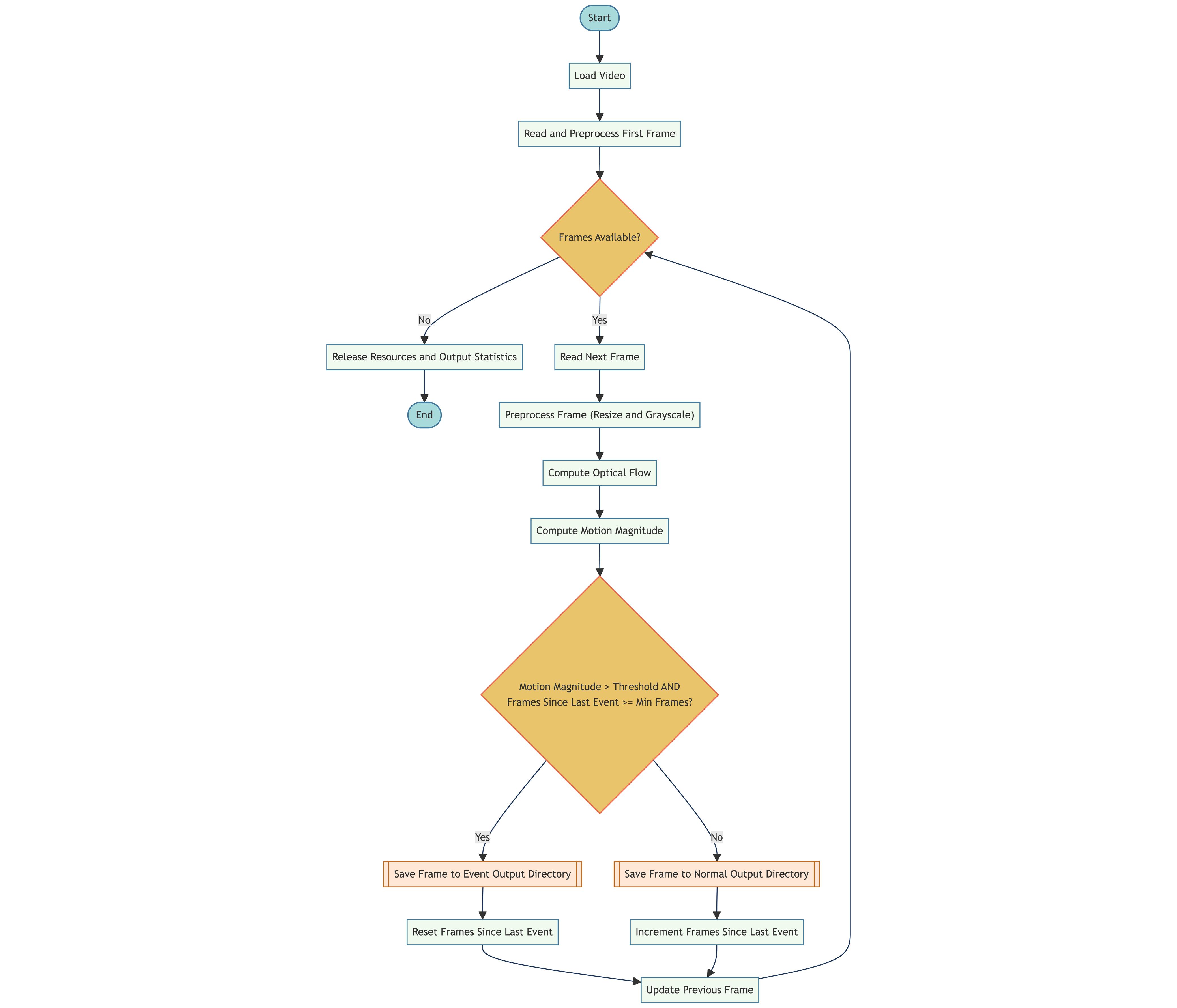} 
    \caption{Video Sampling Approach} 
    \label{fig:sampling} 
\end{figure}

For video frames sampling, we propose an event-based sampling approach for crowd density estimation that selectively captures video frames only when significant changes in crowd movement or density are detected, rather than continuously sampling every frame. This method utilizes optical flow analysis employing the Farneback algorithm to identify frames with substantial motion. Initially, each frame is converted to grayscale to simplify the data and reduce computational complexity. The Farneback algorithm then calculates the optical flow between consecutive frames, assessing pixel displacements in both the x and y directions to capture motion intensity and direction across the frame. From these calculations, the motion magnitude is derived to represent the overall level of movement within the frame. If the average motion magnitude surpasses a predefined threshold, indicating a significant event such as crowd gathering or dispersal, the current frame is saved for further analysis and density map generation, and the previous frame is updated to ensure accurate future motion comparisons. The sampling approach is summarized in Fig \ref{fig:sampling}.

To evaluate the efficiency and effectiveness of our sampling method, we manually labeled a set of ground-truth events and calculated recall, precision, and F1 score, ensuring that significant events were accurately captured. We also computed the reduction ratio to assess storage efficiency, demonstrating the method's capability to significantly decrease data storage requirements. Additionally, we assessed the mean motion magnitudes in both sampled and skipped frames to verify that meaningful events were captured without excessive data loss. For example, in a 49-second video at 24 frames per second, our approach reduced the total frame count to approximately 100–150 frames, effectively decreasing computational load and data storage while preserving critical insights into crowd dynamics. This event-driven sampling is particularly advantageous for applications like security surveillance or sports analytics, where recording only significant events is essential.

\begin{figure}[htbp]
    \centering
    \includegraphics[width=0.5\textwidth]{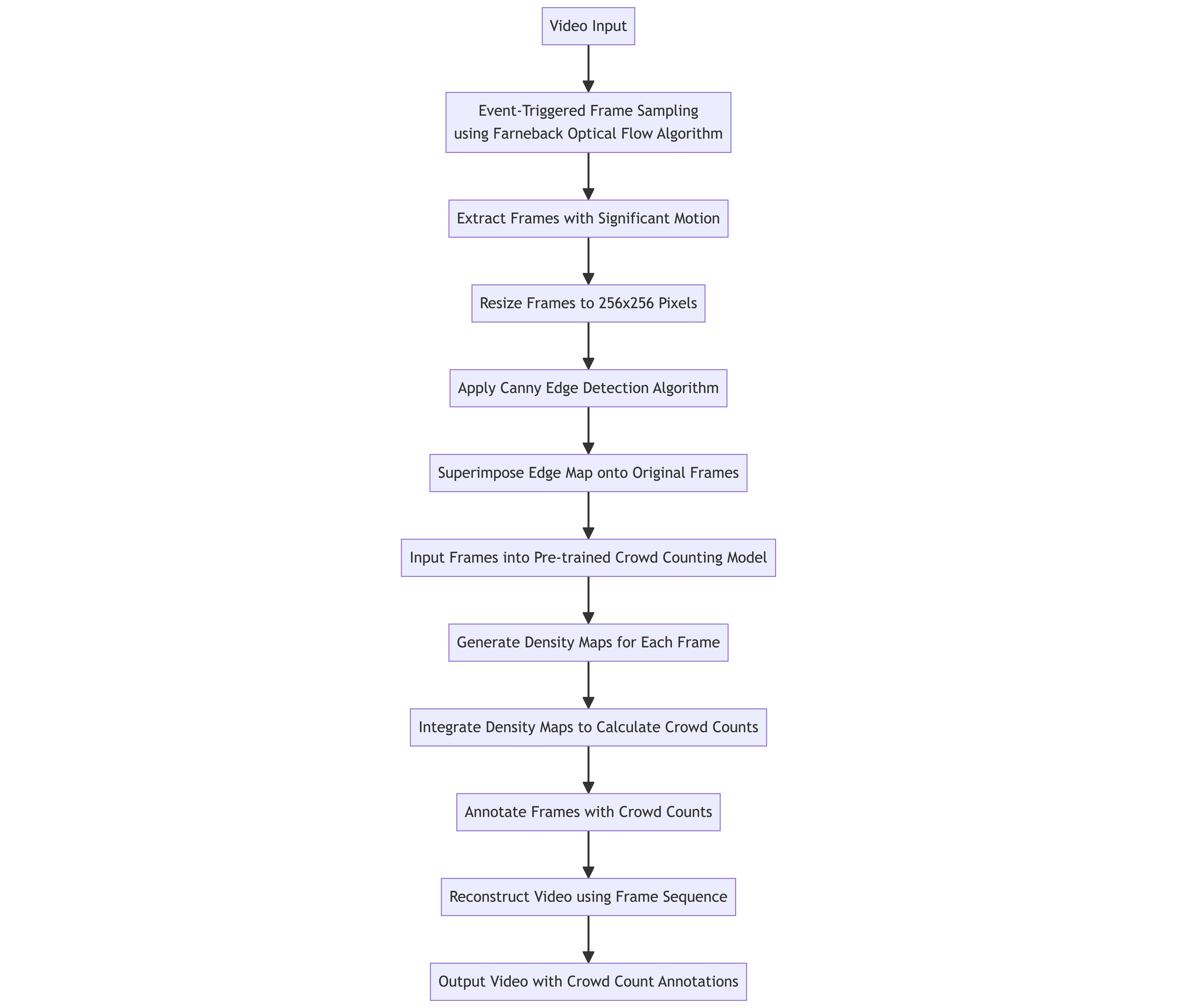} 
    \caption{End-to-end proposed approach} 
    \label{fig:flow} 
\end{figure}

The end-to-end proposed approach is shown in Fig \ref{fig:flow}.

\section{Results}
\label{sec:results}
The qualitative and quantitative analysis was performed on the Shanghai Tech dataset, covering both Part A and Part B. 

\textbf{Our demo for an end-to-end implementation on a sample video is included} \href{https://drive.google.com/file/d/1TmudVHdeCwkDEktFg2n4RwRNmySa4DC9/view?usp=drive_link}{\textbf{here}}.
\subsection{Qualitative Analysis}
\begin{figure}[htbp]
    \centering
    \includegraphics[width=0.5\textwidth]{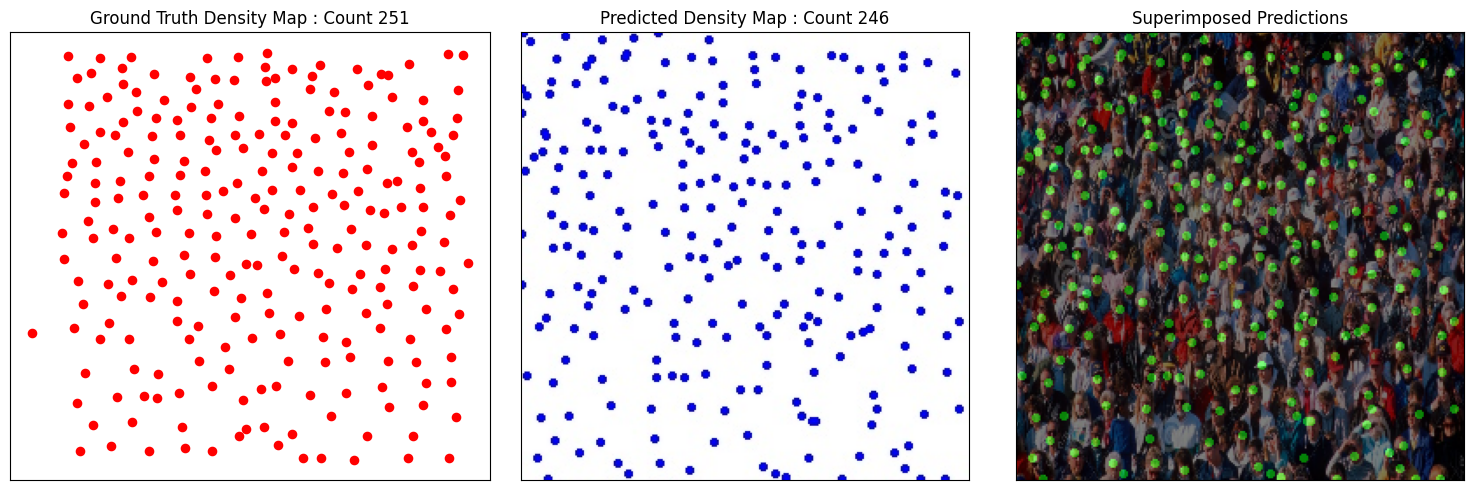} \\[10pt] 
    \includegraphics[width=0.5\textwidth]{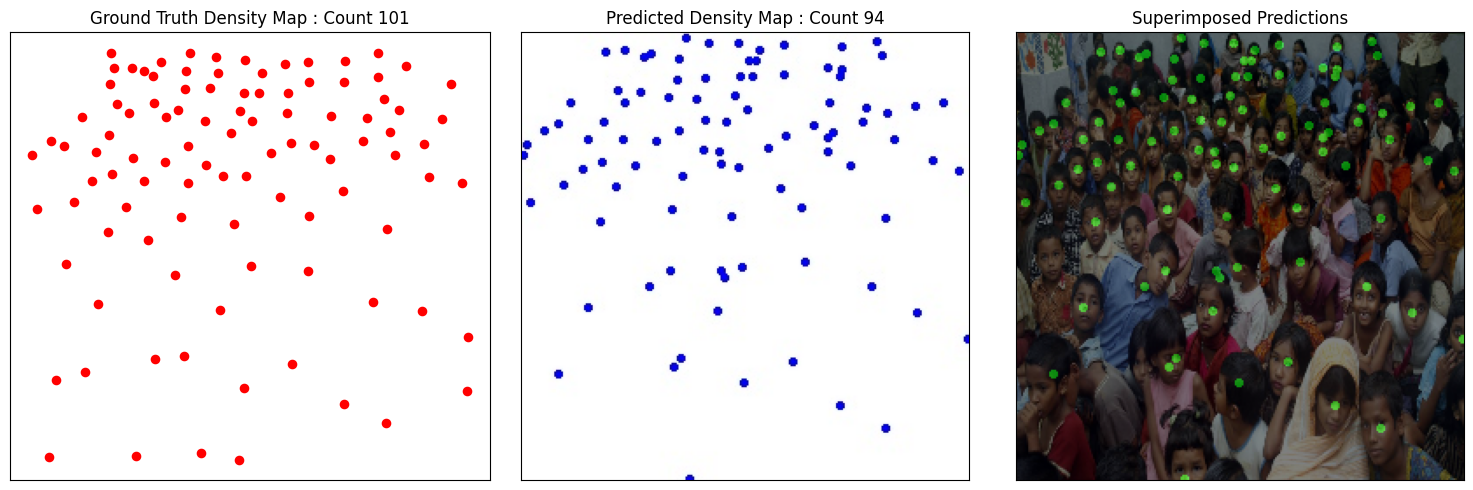} 
    \includegraphics[width=0.5\textwidth]{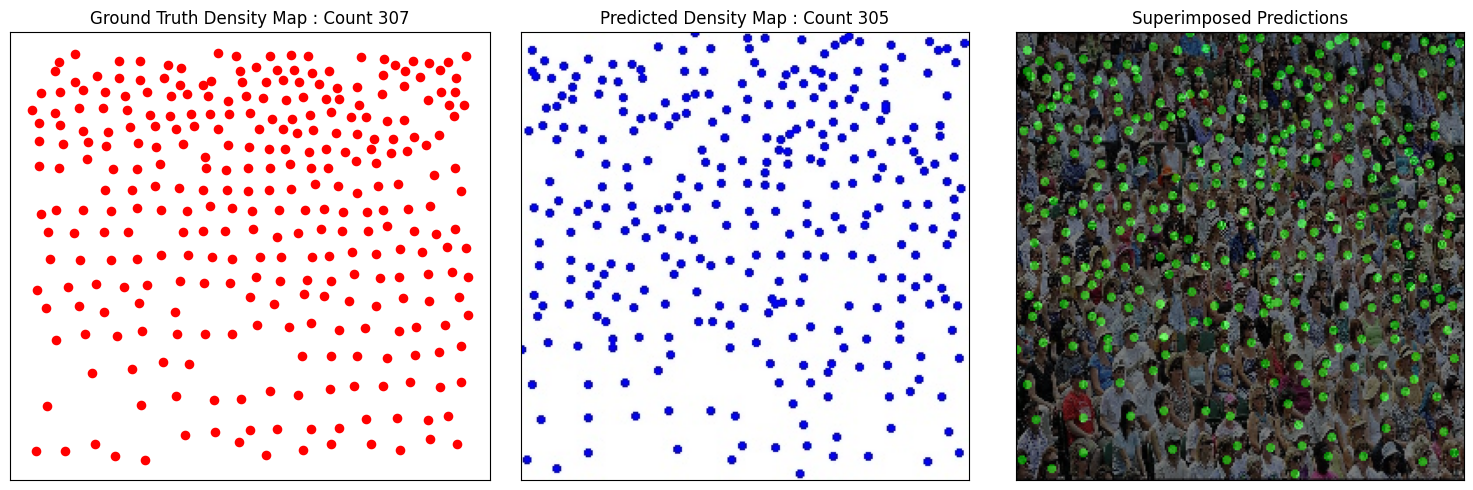} 
    \caption{Part A Samples} 
    \label{fig:samples_a} 
\end{figure}

\begin{figure}[htbp]
    \centering
    \includegraphics[width=0.5\textwidth]{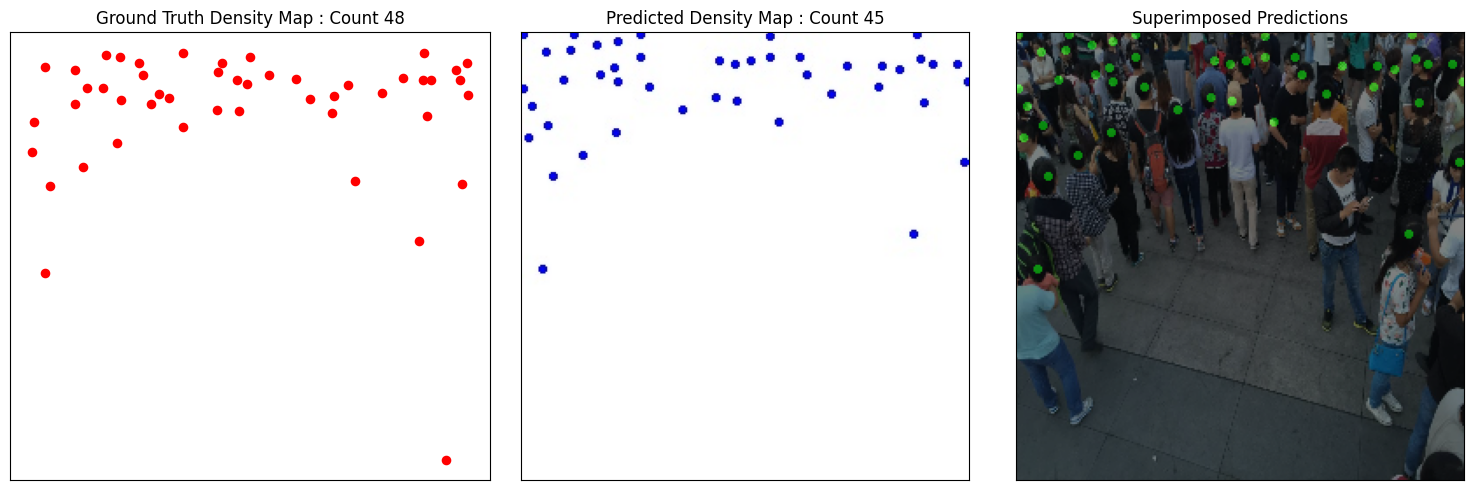} \\[10pt] 
    \includegraphics[width=0.5\textwidth]{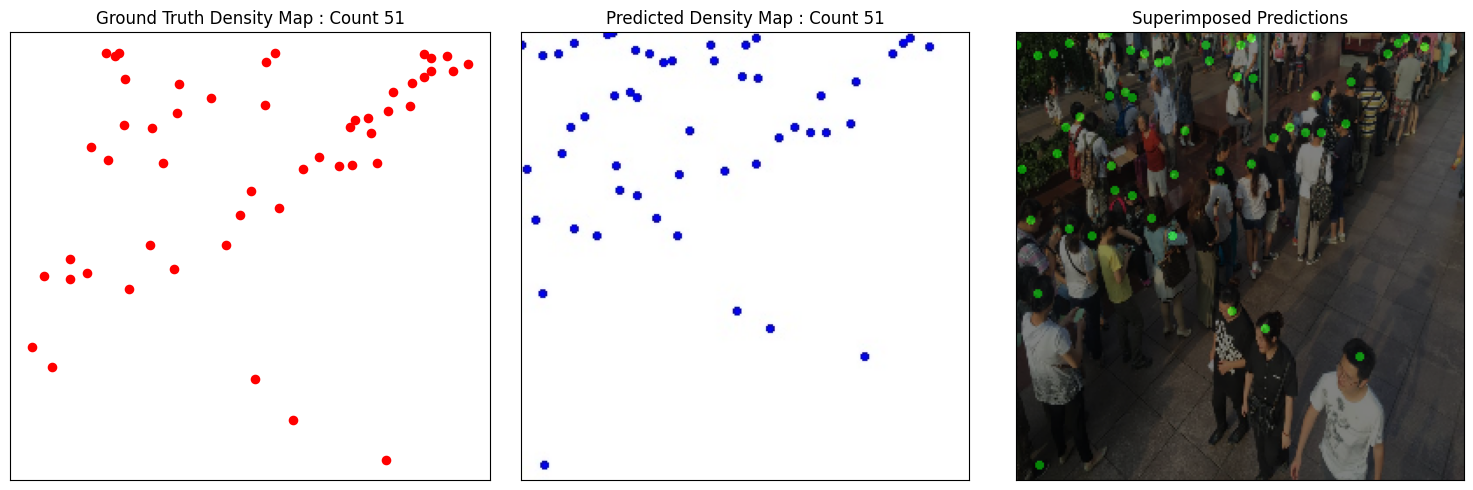} 
    \includegraphics[width=0.5\textwidth]{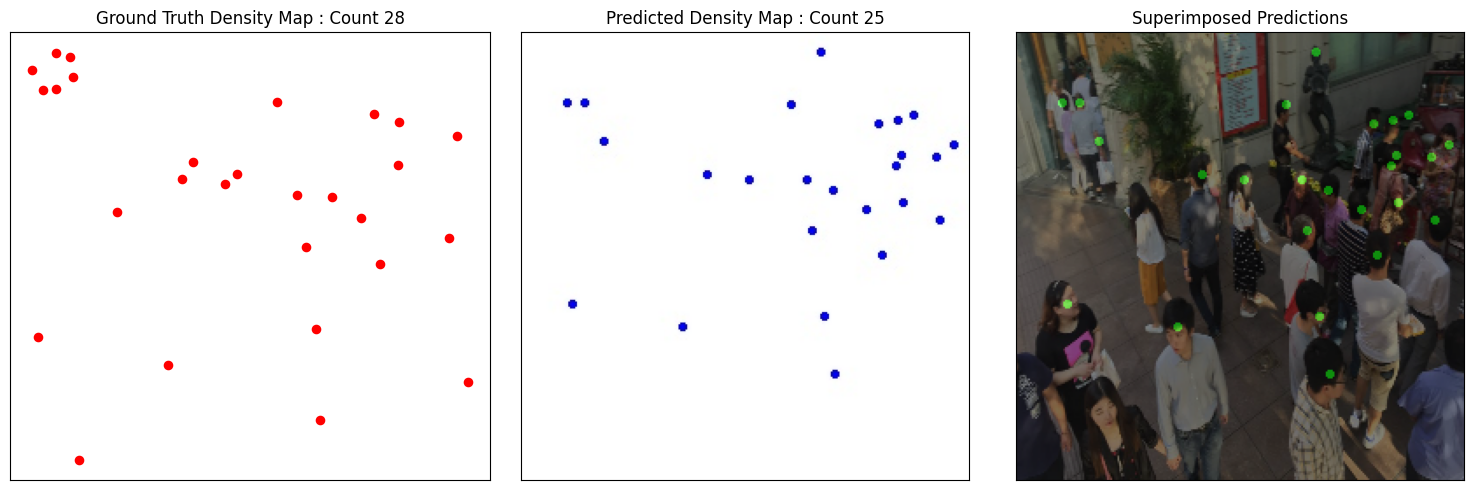} 
    \caption{Part B Samples} 
    \label{fig:samples_b} 
\end{figure}

We evaluate the generated density maps by comparing them with the ground truth maps and overlaying them on the raw images. This highlights the accuracy of the generated density maps. A few qualitative results for both Part A and Part B are presented in Fig \ref{fig:samples_a} and Fig \ref{fig:samples_b} respectively.

\subsection{Quantitative Analysis}
The quantitative evaluation was conducted for both the proposed approach and the video sampling strategy. 

For proposed approach, we compared the MAE with several existing state-of-the-art methods to assess the performance of our approach. It is evident from Table \ref{tab:mae_comparison} that on ShanghaiTech A, the proposed approach achieves an MAE of 54.7, which is competitive compared to other methods. It performs better than methods like TopoCount (61.2) and CLTR (56.9), and is also significantly better than PET (49.3). On ShanghaiTech B, the proposed approach reports an MAE of 6.4, which is among the lowest, performing similarly to methods like CrowdHat (5.7) and STEERER (5.8). It is better than other approaches like TopoCount (7.8).

\begin{table}[htbp]
\centering
\begin{tabular}{|l|c|c|}
\hline
\textbf{Method} & \textbf{ShTech A} & \textbf{ShTech B} \\
\hline
TopoCount & 61.2 & 7.8 \\
SUA & 68.5 & 14.1 \\
ChFl & 57.5 & 6.9 \\
MAN & 56.8 & NA \\
GauNet & 54.8 & 6.2 \\
CLTR & 56.9 & 6.5 \\
CrowdHat & 51.2 & 5.7 \\
STEERER & 54.5 & 5.8 \\
PET & 49.3 & 6.2 \\
CrowdDiff & 47.4 & 5.7 \\
\textbf{Proposed Approach} & \textbf{54.7} &\textbf{6.4} \\
\hline
\end{tabular}
\caption{Comparison of Mean Absolute Error values for different methods on ShanghaiTech A and B datasets.}
\label{tab:mae_comparison}
\end{table}

\begin{table*}[ht]
\centering
\begin{tabular}{|l|c|c|c|c|}
\hline
\textbf{Sampling Method} & \textbf{Saved Frames} & \textbf{Total Frames} & \textbf{Correct Frames} & \textbf{Correct Frame Rate (\%)} \\
\hline
Uniform Sampling & 40 & 1,177 & 15 & 37.5\% \\
Random Sampling & 10 & 1,177 & 4 & 40.0\% \\
Adaptive Sampling & 50 & 1,177 & 28 & 56.0\% \\
Keyframe Sampling & 10 & 1,177 & 6 & 60.0\% \\
Stratified Sampling & 15 & 1,177 & 7 & 46.7\% \\
\textbf{Event-Based Sampling} & \textbf{34} & \textbf{1,177} & \textbf{30} & \textbf{88.2\%} \\
\hline
\end{tabular}
\caption{Comparison of different video sampling methods.}
\label{tab:sampling_methods}
\end{table*}

The Table \ref{tab:sampling_methods} showcases the performance of various video frame sampling methodologies applied to a total of 1,177 frames. Event-Based Sampling proved to be the most effective approach, achieving an impressive 88.2\% correct frame rate by utilizing the Farneback algorithm for optical flow analysis. This method selectively saves frames with significant motion, making it ideal for applications such as security monitoring and sports analysis where capturing high-activity moments is essential. In comparison, other sampling techniques like Adaptive Sampling and Keyframe Sampling demonstrated moderate effectiveness with 56.0\% and 60.0\% correct frame rates, respectively. Stratified Sampling maintained a 46.7\% correct frame rate, ensuring diverse frame selection by proportionally sampling from different video segments. However, methods such as Uniform Sampling and Random Sampling lagged behind, with correct frame rates of 37.5\% and 40.0\%, indicating their lesser ability to identify relevant frames without considering motion or content variability. Overall, the data underscores the superior accuracy of Event-Based Sampling in preserving critical moments, while other methods offer varying levels of effectiveness depending on the application's requirements.

We conducted additional evaluations of our sampling approach on certain videos, calculating key metrics mentioned below and summarizing the results in the Table \ref{tab:metrics}.\\

\begin{itemize}[left=0pt]
    \item \textbf{True Positives (TP)}: The number of correctly identified event frames.
    \item \textbf{False Positives (FP)}: The number of non-event frames incorrectly identified as events.
    \item \textbf{False Negatives (FN)}: The number of event frames that were not identified.
    \item \textbf{Precision}: Measures the accuracy of the positive predictions. Calculated as:
    \[
    \text{Precision} = \frac{\text{TP}}{\text{TP} + \text{FP}}
    \]
    \item \textbf{Recall}: Measures the ability to find all relevant instances. Calculated as:
    \[
    \text{Recall} = \frac{\text{TP}}{\text{TP} + \text{FN}}
    \]
    \item \textbf{F1 Score}: The harmonic mean of Precision and Recall. Calculated as:
    \[
    \text{F1 Score} = 2 \times \frac{\text{Precision} \times \text{Recall}}{\text{Precision} + \text{Recall}}
    \]
\end{itemize}

\begin{table}[h!]
\centering
\begin{tabular}{|l|c|c|c|c|c|c|}
\toprule
\textbf{Video} & \textbf{TP} & \textbf{FP} & \textbf{FN} & \textbf{Precision} & \textbf{Recall} & \textbf{F1} \\
\midrule
Video 1 & 30 & 4 & 5 & 0.882 & 0.857 & 0.869 \\
Video 2 & 9  & 3 & 1 & 0.750 & 0.900 & 0.818 \\
Video 3 & 33 & 7 & 8 & 0.825 & 0.805 & 0.815 \\
Video 4 & 34 & 10 & 8 & 0.773 & 0.810 & 0.791 \\
Video 5 & 54 & 11 & 4 & 0.831 & 0.931 & 0.878 \\
Video 6 & 21 & 2 & 1 & 0.913 & 0.955 & 0.934 \\
\bottomrule
\end{tabular}
\caption{Performance metrics for event-driven sampling across video samples.}
\label{tab:metrics}
\end{table}

From the Table \ref{tab:metrics}, we can conclude that Precision values vary across videos, ranging from 0.750 (Video 2) to 0.913 (Video 6), indicating that the approach performs better in identifying true event frames relative to false positives for some videos.Recall values are generally high, ranging from 0.805 (Video 3) to 0.955 (Video 6), showing the method’s effectiveness in capturing most of the true event frames.The F1 Score, which balances precision and recall, ranges from 0.791 (Video 4) to 0.934 (Video 6). This suggests that the sampling approach achieves robust performance across different video samples.

\section*{Conclusion}

In this work, we successfully extended a crowd density estimation approach from static images to videos, leveraging conditional diffusion models to address the challenges of temporal dynamics and computational efficiency. The proposed models generated high-fidelity density maps that accurately captured crowd distributions in sampled frames, with estimated crowd counts closely aligning with ground-truth data. To tackle the additional complexity of video processing, we introduced an event-driven sampling strategy based on the Farneback optical flow algorithm. This approach effectively reduced computational overhead while preserving critical crowd dynamics, achieving significant reductions in frame processing without compromising accuracy. 
The event-based sampling method demonstrated strong performance, highlighting its ability to identify significant frames with high recall and precision. This efficiency is particularly valuable for real-time crowd monitoring applications, where constraints on processing speed and storage capacity are paramount. The advancements presented in this study contribute to the development of scalable, accurate, and efficient frameworks for monitoring crowd behavior in dynamic and high-stakes scenarios such as public safety, disaster response, and event management.